\documentclass[lettersize,journal]{IEEEtran}

\usepackage[utf8]{inputenc}
\usepackage[T1]{fontenc}
\usepackage{amsmath,amsfonts,amssymb,bm}
\usepackage{algorithmic}
\usepackage{algorithm}
\usepackage{array}
\usepackage[caption=false,font=normalsize,labelfont=sf,textfont=sf]{subfig}
\usepackage{textcomp}
\usepackage{stfloats}
\usepackage{url}
\usepackage{verbatim}
\usepackage{graphicx}
\usepackage{cite}
\usepackage[dvipsnames]{xcolor}
\usepackage{booktabs}
\usepackage{multirow}
\usepackage{tabularx}
\usepackage{threeparttable}
\usepackage{paralist}
\usepackage{pifont}
\usepackage[accsupp]{axessibility}
\usepackage{microtype}
\usepackage{nicefrac}
\usepackage{capt-of}
\usepackage{cuted}
\graphicspath{{figures/}{figures_appendix/}}
\newcommand{\NOTE}[1]{\textcolor{red}{}}

\newcommand{\CUT}[1]{}

\definecolor{MyGreen}{RGB}{0,128,0}
\definecolor{MyRed}{RGB}{220,20,60}

\usepackage{amsmath,amsfonts,bm}

\def\eqref#1{equation~\ref{#1}}

\def\1{\bm{1}}

\DeclareMathAlphabet{\mathsfit}{\encodingdefault}{\sfdefault}{m}{sl}
\SetMathAlphabet{\mathsfit}{bold}{\encodingdefault}{\sfdefault}{bx}{n}

\begin{document}

\title{StreetDiff: Multi-view Street Scenes Generation via Cross-view Consistent Multi-view Stable Diffusion with Structure Prompts}

\author{Qi~Zhang,~\IEEEmembership{Member,~IEEE,}
        Yanyifan~Wang, 
        Weiyuan~Zhang, 
        and~Hui~Huang*,~\IEEEmembership{Senior Member,~IEEE}
\thanks{Qi Zhang, Yanyifan Wang, Weiyuan Zhang, and Hui Huang are with the College of Computer Science and Software Engineering, Shenzhen University, China. 
E-mail: qi.zhang.opt@gmail.com, \{2400101089, 2400101002\}@mails.szu.edu.cn, hhzhiyan@gmail.com}
\thanks{*Corresponding author.}
\thanks{Manuscript received xxx; revised August xxx.}
}

\markboth{IEEE Transactions on Multimedia,~Vol.~xx, No.~xx, 2026}%
{Qi Zhang \MakeLowercase{\textit{et al.}}: StreetDiff}

\maketitle

\begin{strip}
\centering
\vspace{-2.5cm}
\includegraphics[width=\textwidth]{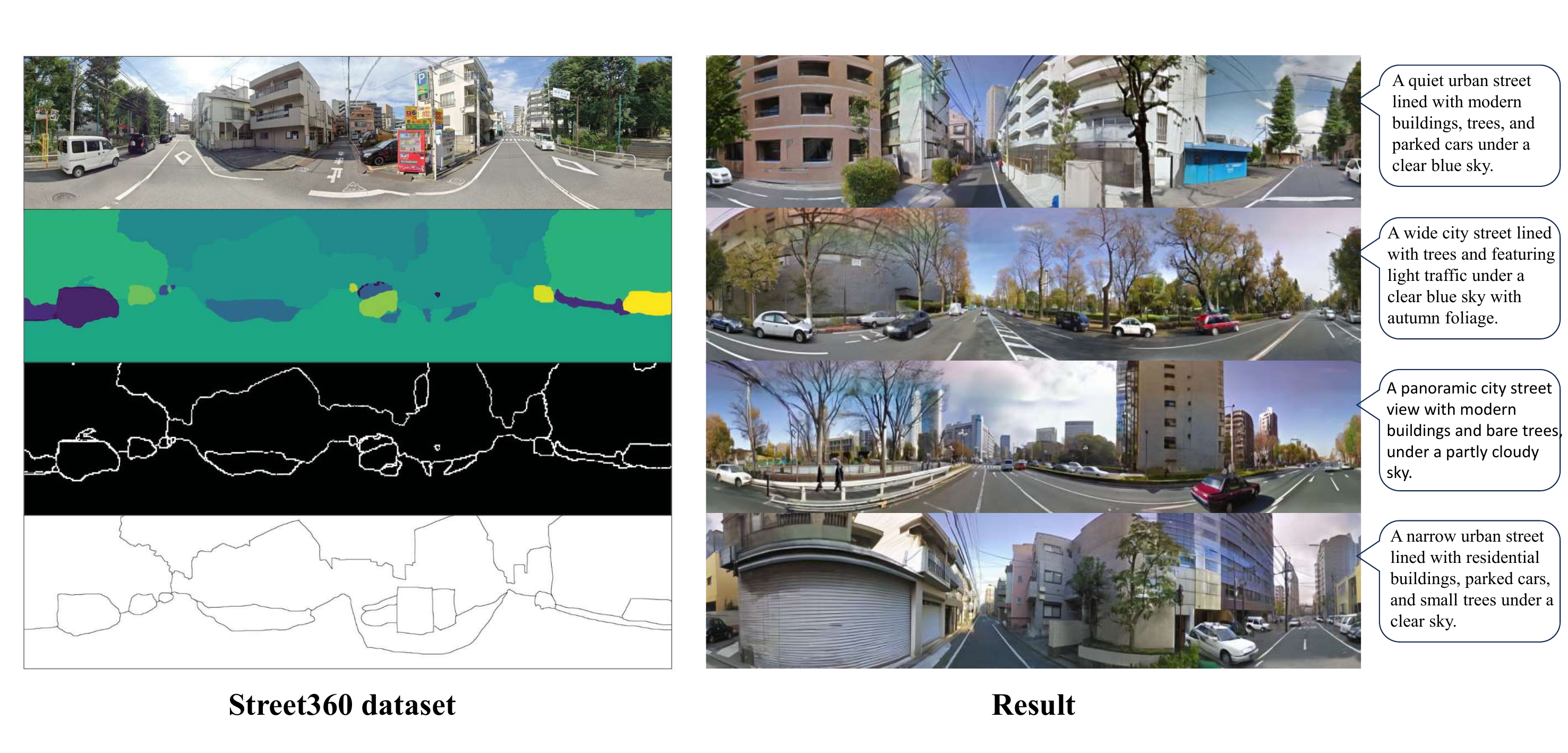}

\captionof{figure}{Representative samples and structure prompts from the proposed Street360 dataset are shown on the left, and representative results generated by StreetDiff are shown on the right. The examples on the two sides are independently selected to illustrate the dataset characteristics and the model's generation ability.}
\vspace{-0.2cm}
\label{fig:dataset}
\end{strip}

\begin{abstract}


Multi-view diffusion models have shown strong performance in scenes with strong geometric priors and sparse semantics, such as indoor rooms or simple outdoor environments (e.g., fields, courtyards). However, they often fail to maintain cross-view consistency under camera rotation, especially in structurally complex urban environments. Without explicit modeling of spherical correspondence across views, existing approaches tend to produce object duplication, structural distortion, and layout inconsistency. To address this limitation, we propose StreetDiff, a multi-view diffusion framework that explicitly enforces cross-view alignment during denoising. StreetDiff introduces a Panorama--Perspective Synergy design to decouple global layout reasoning from local detail synthesis, and incorporates a Panorama Alignment Module (PAM) that establishes spherical-projection-based attention constraints across views. By injecting structured alignment constraints without modifying the diffusion backbone, our framework achieves robust cross-view coherence in challenging urban street scene generation tasks. In addition, we construct Street360, a large-scale HDR multi-view urban panorama dataset. Extensive experiments demonstrate that StreetDiff significantly improves structural consistency and visual fidelity compared to prior multi-view diffusion generation methods.

\end{abstract}

\section{Introduction}\label{sec:intro}

Text-based multi-view image generation \cite{ramesh2021zeroshottexttoimagegeneration,nichol2022glidephotorealisticimagegeneration,ramesh2022hierarchicaltextconditionalimagegeneration,Rombach_2022_CVPR,NIPS_Photorealistic_t2i} has emerged as a promising direction in computer vision with applications in VR\cite{yang2023dreamspacedreamingroomspace}, AR, video games, and film production. Early methods \cite{wang2023360,BarTal2023MultiDiffusionFD,Lee2023SyncDiffusionCM,PanoFree} extended pretrained diffusion models to panoramic generation by iteratively synthesizing local views, but were mainly constrained to indoor scenes. Later approaches \cite{panfusion2024,chen2022text2light,MVDiffusion} fine-tuned diffusion models to directly generate 2:1 panoramas, yet their resolution remained limited by the original architectures (e.g., 1024$\times$512). As a result, outdoor panoramas, particularly street scenes, remain challenging due to their complex layouts and dynamic elements.
In summary, two challenges hinder progress: (\textit{i}) most existing datasets \cite{dai2017scannet,Matterport3D,chen2022text2light,xiao2012sun360} emphasize indoor or overly simple natural scenes, lacking urban complexity; (\textit{ii}) current methods often produce artifacts such as distorted buildings and warped road structures when applied to street scenes. 

To bridge this gap, we introduce Street360, a large-scale HDR multi-view urban panorama dataset with aligned text and dense structural annotations, as shown in Fig.~\ref{fig:dataset}. Leveraging this resource, we propose StreetDiff, a panorama--perspective synergy framework that uses a global panoramic representation and a geometry-aware Panorama Alignment Module (PAM) to enforce spherical cross-view consistency during denoising. Extensive experiments demonstrate that our approach significantly improves visual fidelity and multi-view geometric coherence compared to state-of-the-art baselines, particularly in text-only generation scenarios.

In summary, the contributions of the paper are as follows.
\begin{compactitem}
    \item We introduce \textbf{Street360}, a large-scale HDR multi-view street panorama dataset with aligned text prompts and dense structural annotations, designed to support realistic urban street scene generation and evaluation.
    \item We propose a novel Panorama--Perspective Synergy
    Framework, \textbf{StreetDiff}, which effectively leverages structure prompts (\textit{e.g.}, segmentation maps, contour maps, or user sketches) to guide the generation, ultimately producing street images that accurately adhere to the given structural prompts. Furthermore, to exploit global contextual information and enforce consistency across multiple views, we introduce the \textbf{Panorama Alignment Module (PAM)}, which establishes geometry-aware correspondences between panoramic and perspective representations. 

    \item Extensive experimental results demonstrate our method achieves the best performance in street scene generation, surpassing previous models in terms of generation quality and multi-view consistency. 
\end{compactitem}

\section{Related Work}
\label{sec:relatedwork}

\textbf{Panoramic Image Generation.}
Several methods \cite{360-Degree_Panorama_Generation_from_Few_Unregistered_NFoV_Images,GAN360,360-Degree_Outpainting,chen2022text2light,panfusion2024,StyleLight,BIPS,Customizing_360-Degree_Panorama_Diffusion_Models,Sun_2025_CVPR,zheng2025panorama,yang2024layerpano3d} have been proposed for panoramic scene generation. \textcolor{black}{These works either progressively outpaint panoramas in an autoregressive way, or generate the entire panorama in one shot.} 
Other methods, such as MultiDiffusion \cite{BarTal2023MultiDiffusionFD} and SyncDiffusion \cite{Lee2023SyncDiffusionCM}, divide the panorama into multiple regions, generating them step by step to address inconsistencies between regions.
PanFusion \cite{panfusion2024} takes this step further by integrating a panoramic branch and a multi-view branch, leveraging their projection relationships to achieve more realistic panorama generation, \textcolor{black}{but the generated images are of relatively low resolution.} 
\textcolor{black}{OPa-Ma \cite{gao2024opamatextguidedmamba} uses lightweight Mamba to model long-range dependencies in panoramas and leverages text for better semantic consistency. }\textit{However, these methods are mainly validated in indoor or simple scenes and often struggle to preserve global structure in unbounded urban streets. The absence of a global geometric scaffold leads to severe distortions in buildings and roads when applied to unbounded, layout-prior-weak environments.}


\textbf{Multi-view Generation.}
While single-view image generation \cite{ddpm,DiffusionBeatGAN,NIPS_Photorealistic_t2i} has become relatively mature, multi-view image generation still faces several challenges. 
For example, MVDiffusion \cite{MVDiffusion} maintains consistency between adjacent views using the Correspondence-Aware Attention (CAA) module; however, issues such as duplicated objects and repetitive textures may still arise. 
\textcolor{black}{DiffCollage \cite{zhange2023diffcollage} decomposes large images into factor graphs and generates them with multiple diffusion models in parallel, but struggles with tasks requiring long-range consistency, often producing artifacts such as duplicated structures (e.g., repeated snake tails).}
In the field of 3D generation, some methods \cite{liu2023one2345,liu2023zero1to3,tang2024mvdiffusionpp,latent_nerf}, such as Zero-1-to-3 \cite{liu2023zero1to3}, leverage geometric relationships across multiple views to predict a complete 3D representation, while Latent-NeRF \cite{latent_nerf} integrates NeRF as a 3D prior and applies diffusion in the latent space, resulting in more natural and consistent multi-view generation. 
These approaches 
provide valuable insights into multi-view diffusion models, yet improving view consistency and geometric accuracy remains an open challenge. 
\textit{To address this, we propose a Panorama--Perspective Synergy Framework based on structural information. The structural information enhances the geometric stability of the generated images, while the Panorama--Perspective Synergy Framework integrates global and local information to further improve consistency, effectively mitigating instability and repetitive texture issues in multi-view generation.}

\textbf{Outdoor Generation.}

Although existing methods \cite{chen2022text2light,360-Degree_Outpainting,BarTal2023MultiDiffusionFD,Lee2023SyncDiffusionCM,MVDiffusion,panfusion2024,PanoFree} achieve strong performance on indoor scenes, outdoor generation remains challenging due to higher structural and semantic complexity. In practical applications such as autonomous driving, realistic and coherent outdoor multi-view generation is essential. DrivingDiffusion \cite{DrivingDiffusion} and PERLDIFF \cite{zhang2024perldiff} introduce 3D semantic boxes for structured synthesis, while PanoFree \cite{PanoFree} and MVDiffusion \cite{MVDiffusion} extend diffusion models to certain outdoor environments, though often limited to relatively simple scenes. Mixed-View Panorama Synthesis \cite{xiong2024mixed} further combines satellite imagery and nearby street views with geospatial attention to generate target panoramas. Recent cross-view approaches, including CrossViewDiff \cite{li2024crossviewdiffcrossviewdiffusionmodel}, SkyDiffusion \cite{ye2024skydiffusion}, Controllable Satellite-to-Street-View Synthesis \cite{ze2025controllable}, and Geometry-Guided Cross-View Diffusion \cite{Geometry-Guided-Cross-View-Diffusion}, leverage BEV representations, homography adjustment, or geometry-guided conditions to mitigate viewpoint gaps and geometric ambiguity. In addition, Streetscapes \cite{deng2024streetscapes} and StreetCrafter \cite{yan2024streetcrafter} explore controllable street video generation using structural cues such as LiDAR-rendered point clouds. These methods demonstrate progress in outdoor or cross-view synthesis but do not explicitly address consistent multi-view generation in complex urban street environments. Compared to these approaches, this paper focuses on generating urban street scenes, achieving detailed and realistic renderings of key elements such as buildings, vehicles, and crosswalks. \textit{To this end, our proposed StreetDiff, a Panorama--Perspective Synergy Framework combining panoramic and multi-view generation, introduces structure guidance to provide a comprehensive solution for text-driven multi-view image and 360 panorama generation.}


\section{StreetDiff Model}
\label{sec:method}

\begin{figure*}[t]
    \centering
    \includegraphics[width=\linewidth]{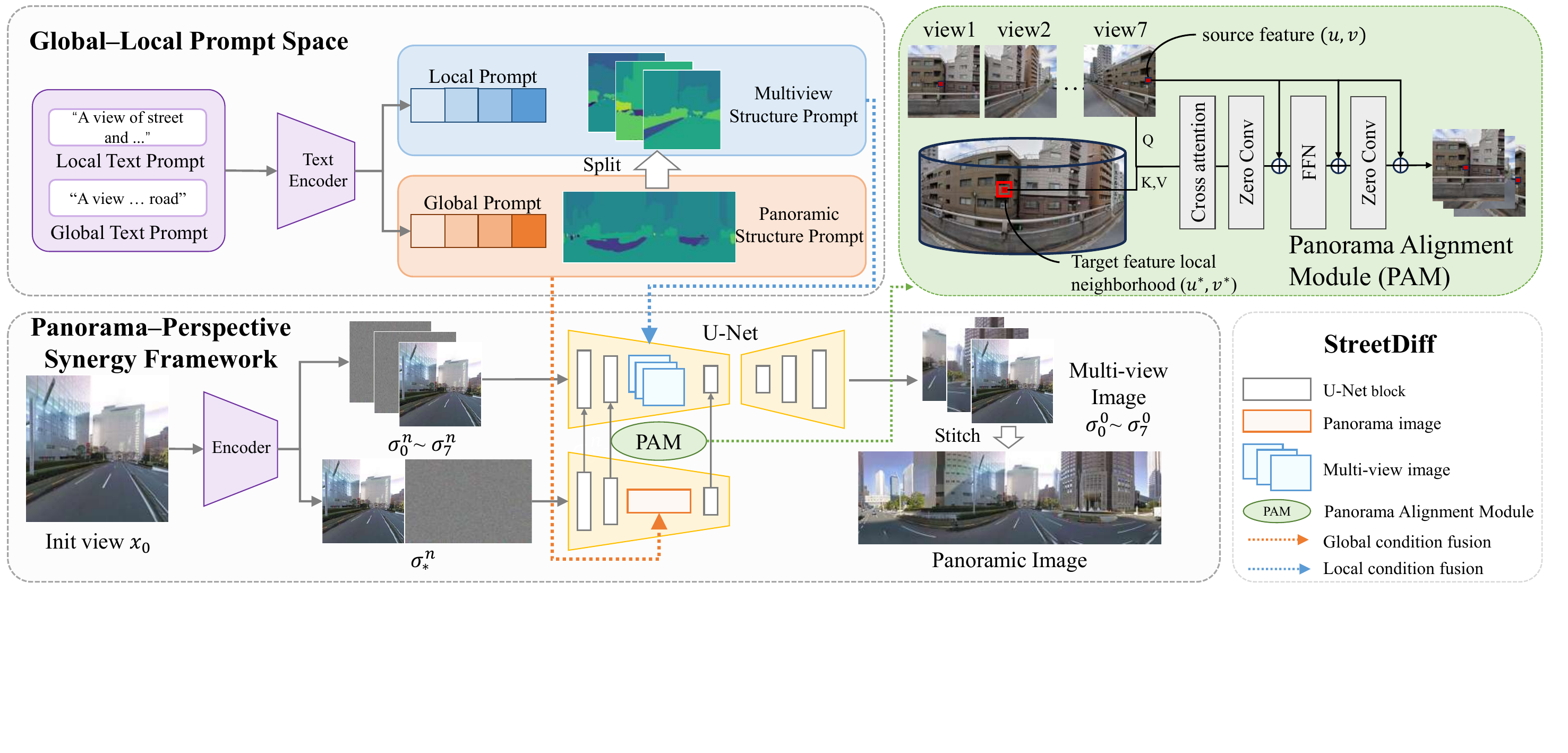}
    \vspace{-2.4cm}
    \caption{The pipeline of StreetDiff. The top is the Global-Local Prompt Space, including local and global text prompts, and the corresponding structure prompts. The global text is composed by concatenating the local texts, while the multi-view structure prompt is derived by segmenting the panoramic structure prompt. The bottom is a Panorama--Perspective Synergy Framework: a panoramic generation and a multi-view generation module. The global and local conditions are inserted into the panorama and multi-view generation branches, respectively.}
    \label{fig:pipeline}
    \vspace{-0.4cm}
\end{figure*}

The goal of the proposed StreetDiff model is to generate coherent multi-view street images, aligned with user intent, restricted to non-polar regions to avoid distortions, given a single target street view and its corresponding structural prompts. The structural prompts can take various forms, such as contour maps, semantic maps, or user-drawn sketches.
As illustrated in Fig.~\ref{fig:pipeline}, our \textbf{Global--Local Prompt Space} extracts both local and global textual descriptions as well as contour-based structural prompts (Sec.~\ref{subsec:prompt}). These serve as inputs to the \textbf{Panorama--Perspective Synergy Framework}, which generates eight perspective views that are subsequently stitched into a complete panorama (Sec.~\ref{subsec:framework}). To effectively leverage global information, we propose the \textbf{Panorama Alignment Module (PAM)}, which maps each point from a perspective view to a local neighborhood in the panorama, thereby ensuring cross-view consistency (Sec.~\ref{subsec:PAM}). Finally, to further enhance the quality of the \textbf{Panorama--Perspective Synergy Framework}, we conduct separate training for the panoramic and perspective modules, followed by joint training with PAM (Sec.~\ref{subsec:training}).

\subsection{Global--Local Prompt Space}
\label{subsec:prompt}
As the first stage of our pipeline, we construct a unified prompt space that bridges panoramic images and their corresponding multi-view representations. This prompt space integrates both \textbf{textual} and \textbf{structural} conditions, capturing scene semantics at different levels of granularity. 

\textbf{Text Prompts.} StreetDiff combines local and global text prompts. Specifically, we first realize \cite{Equirec2Perspec} a panorama by generating eight perspective views, each possessing a horizontal field of view of 90$^\circ$ with a 45$^\circ$ overlap and employ BLIP-3 \cite{xue2024xgenmmblip3familyopen} to generate initial local text prompts for each view. However, adjacent views often yield highly similar descriptions, which limits their discriminative power. To address this, we introduce GPT-4 to regenerate more detailed local prompts for each perspective view by leveraging both the panoramic image and the BLIP-3 outputs. Finally, all refined local prompts are concatenated to form a global text prompt, providing a more comprehensive and semantically rich scene description. 
All refined local prompts are used as the multi-view text inputs for MVDiffusion and for the multi-view module of our method, while the concatenated global prompt serves as the text input for panoramic baselines and for the panoramic module of our model. The semantic information contained in both forms is derived from the same content, and the distinction lies only in whether the prompts are supplied separately or as a single merged description.


\par
\textbf{Structure Prompts.} We use three types of structure prompts (see Fig.~\ref{fig:dataset}) in the StreetDiff model: segmentation maps, contour maps, and user-input sketches.
Panoramic segmentation maps are obtained using OneFormer \cite{jain2023oneformer}, and further divided into multi-view structure prompts via the same viewpoint cropping as in text prompts \cite{Equirec2Perspec}.
The contour maps are extracted as boundaries from the segmentation maps, while the sketches are user-drawn depictions of street scenes.
These structure prompts provide important guidance for generating fine structures in the street scenes, such as buildings, roads, \textit{etc.}

\begin{figure*}[t]
    \centering
    \includegraphics[width=1.0\linewidth]{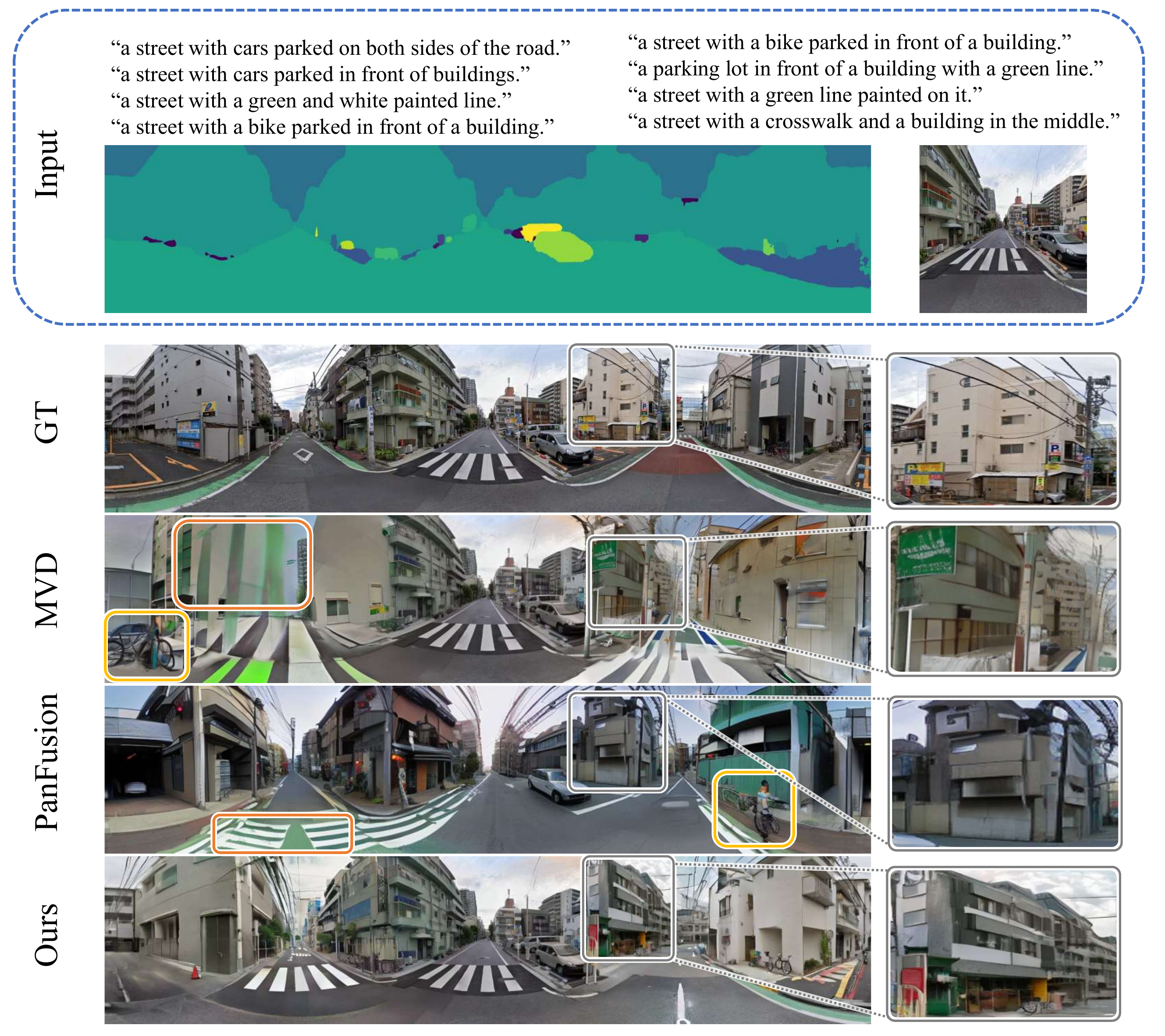}
    \vspace{-0.6cm}
    \caption{Qualitative comparisons of panorama generation. We present the results obtained by stitching together multi-view images generated by MVDiffusion and StreetDiff. Using color-coded boxes, we highlight issues such as \textcolor[RGB]{237, 125, 49}{distorted lines} and \textcolor[RGB]{255, 192, 0}{unrealistic generated items} found in MVDiffusion \cite{MVDiffusion} and PanFusion \cite{panfusion2024}, which are effectively addressed by our method. }
    \label{fig:sem_genertaion}
\end{figure*}

\subsection{Panorama--Perspective Synergy Framework}
\label{subsec:framework}

To address the cross-view inconsistency that is especially severe in complex street scenes, we design a \textbf{two-module framework} consisting of a \textit{panoramic module} and a \textit{multi-view module}, both adapted from the pretrained SD U-Net (see Fig.~\ref{fig:pipeline}). Unlike PanFusion~\cite{panfusion2024}, which aggregates multi-view features in a bottom-up manner to synthesize a panorama, our framework inverts this paradigm: we leverage a global panoramic representation as a top-down global prior to explicitly constrain and guide the generation of consistent perspective views. Both modules collaborate during the diffusion denoising process, optimizing the multi-view latent representation that is decoded into the multi-view outputs.

Unlike PanFusion~\cite{panfusion2024}, which aggregates multi-view features in a bottom-up manner to synthesize a panorama, our framework inverts this paradigm: we leverage a global panoramic representation as a top-down global prior to explicitly constrain and guide the generation of consistent perspective views.
Both modules collaborate during the diffusion denoising process, optimizing the multi-view latent representation that is decoded into the multi-view outputs.  

\textbf{Multi-view Module.}  
Following MVDiffusion \cite{MVDiffusion}, each generated perspective view has a 90$^\circ$ field of view (FOV) with 45$^\circ$ overlap. This module operates in standard pinhole camera projection space and focuses on synthesizing high-resolution local textures at $512 \times 512$ resolution and realistic view-dependent details. A pretrained multi-view SD backbone is employed, and structural cues derived from the panoramic module are injected into the U-Net features to improve inter-view consistency and reduce structural distortion. We further finetune this module on our dataset to better capture domain-specific characteristics before joint training with the panoramic module.

\textbf{Panoramic Module.}  
The panoramic module synthesizes a $2048 \times 512$ representation in spherical projection space. This module captures holistic scene layout, long-range structural continuity, and global spatial priors that are difficult to model purely from perspective views. We leverage LoRA \cite{hu2022lora} for efficient adaptation and apply rotation augmentation with cyclic padding during denoising to preserve loop consistency. Crucially, unlike PanFusion, this module does not produce a final image; its output serves solely as a structural prior to guide multi-view synthesis via the PAM. By conditioning the diffusion on panoramic structural information, we improve both the quality of the panoramas and their ability to regularize the multi-view synthesis.  \par

\subsection{Model Training}
\label{subsec:training}

\subsection{Panorama Alignment Module (PAM)}
\label{subsec:PAM}



Since the panoramic and perspective modules operate under different projections, cross-branch interaction requires projection-consistent alignment. Under camera rotation, a perspective pixel corresponds to a projected region on the panorama rather than an exact point. We therefore introduce PAM to enforce spherical-consistent alignment and prevent mismatched fusion that breaks loop closure.

Specifically, given a pixel \((u,v)\) in a perspective view, we first back-project it to normalized camera coordinates as \(\tilde{x}_p = K^{-1}[u,v,1]^\top\), where \(K \in \mathbb{R}^{3\times 3}\) denotes the intrinsic matrix. The direction in world space is then obtained by applying the camera rotation \(R \in SO(3)\) and normalization, i.e., \(d = \tfrac{R\tilde{x}_p}{\lVert R\tilde{x}_p \rVert}\). The direction vector \(d = (d_x,d_y,d_z)^\top\) is further converted to spherical angles \(\theta = \arctan2(d_x,d_z)\) and \(\phi = \arcsin(d_y)\), and mapped to panorama coordinates \((u_e,v_e)\) of resolution \((W_e,H_e)\) as \(u_e = \tfrac{\theta + \pi}{2\pi} W_e\) and \(v_e = \tfrac{\pi/2 - \phi}{\pi} H_e\).\par


This step establishes a \textbf{projection-consistent correspondence anchor}, which defines the geometrically aligned location between perspective and panoramic representations. The inverse mapping (from panorama to perspective) can be derived analogously by converting $(u_e,v_e)$ to a spherical direction vector and projecting it back with $(K,R)$. In the feature space, PAM avoids dense warping and instead leverages the above mapping as an anchor for cross-branch attention. Concretely, a query token $q$ from the target branch is projected to the corresponding position $(u^*,v^*)$ in the source branch, and cross-attention is performed within its local neighborhood $N(u^*,v^*)$:
\begin{equation}
\text{Attn}(q) = \text{Softmax}\!\left(\frac{Q_qK_N^\top}{\sqrt{D}} + B_N \right)V_N, \tag{1}
\end{equation}
where $Q_q \in \mathbb{R}^D$ is the query vector, $K_N, V_N \in \mathbb{R}^{|N|\times d}$ are the sampled key and value features, and $B_N$ is a Gaussian bias emphasizing the projected center.  \par
To adapt PAM to the multi-scale design of the U-Net, we vary the neighborhood size across layers: high-resolution layers (\(64 \times 64, 32 \times 32\)) adopt small radii (\(r=1,2\), corresponding to \(3 \times 3\) and \(5 \times 5\) windows) to preserve local details; the intermediate layer (\(16 \times 16\)) uses a larger radius (\(r=3\), corresponding to a \(7 \times 7\) window) for broader context; 
and the bottleneck (\(8 \times 8\) or \(4 \times 4\)) directly attends to the entire feature map. This schedule ensures a balance between fine-grained detail preservation and global geometric consistency.

The attention result is further processed by a zero-initialized $1\times 1$ convolution and fused with the original features in a residual manner:
\begin{equation}
F_{\text{out}} = F_{\text{in}} + \text{Conv}^{(0)}_{1\times 1}(\text{Attn}(q)), \tag{2}
\end{equation}
where $F_{\text{in}}, F_{\text{out}} \in \mathbb{R}^{c\times h\times w}$ denote the input and output features.  

By integrating spherical projection with cross-attention, PAM enforces geometry-aware alignment while avoiding artifacts introduced by dense feature warping. When applied at multiple scales of the U-Net, PAM not only preserves the generative capacity of the diffusion backbone, 
but also significantly improves the consistency between panorama and perspective branches.



Unlike the one-stage training methods MVDiffusion \cite{MVDiffusion} and PanFusion \cite{panfusion2024}, our StreetDiff adopts a three-stage training strategy, with a particular emphasis on the incorporation of structural information to enhance the quality of the generated results. 
Our StreetDiff borrows the dual-branch architecture from PanFusion and the CAA from MVDiffusion, but differs by introducing panoramic and multi-view structural prompts as conditional inputs to guide the model's spatial layout and geometric consistency during image generation. Additionally, we use the panoramic module as an explicit structural prior to guide multi-view generation. The training process is: 
\begin{compactitem}
\item  \textbf{Stage 1}: We first fine-tune SD on the Street360 dataset to acquire outdoor street scene priors. At this stage, we conduct two types of training on the single-view U-Net: one focuses on single-view street image generation under a normal perspective (FOV=90$^\circ$, aspect ratio 1:1) to learn local structural priors of street scenes, while the other directly generates panoramic images (aspect ratio 4:1) to learn distribution characteristics under panoramic projection. In this way, the model learns both panoramic and view-based structural priors, enabling generation of single-view or coarse panoramic images that are structurally plausible per view but lack cross-view consistency.

\item \textbf{Stage 2}: We train the multi-view module to enforce consistency among its generated views by modeling inter-view geometric dependencies during denoising. This is achieved through an attention-based mechanism that aligns features across perspectives using shared structural cues, without requiring external panoramic guidance.

\item \textbf{Stage 3}: We jointly fine-tune PAM and both the panoramic and multi-view modules to strengthen spherical cross-view consistency. PAM learns geometry-aware correspondences that propagate the panoramic structural prior to perspective-view synthesis, aligning local generation with the global spherical layout. This yields 360$^\circ$ azimuth-consistent multi-view outputs while retaining the visual fidelity learned in Stages 1--2.
\end{compactitem}
For the multi-view noises ($\epsilon_0 \sim \epsilon_7$) and the panoramic noise ($\epsilon^*$), we employ the same latent map initialization strategy so that the noise at spatially aligned positions between the multi-view images and the panorama remains consistent. In Stage 1 and Stage 2, given the panoramic image $x^*$ and the multi-view images ($x_0 \sim x_7$) obtained by cropping $x^*$, the losses for the panoramic module $\mathcal{L}^*$ and the multi-view module $\mathcal{L}_i$  are formulated as follows.

\begin{equation}
\mathcal{L}^* = \mathbb{E}_{\mathcal{E}(x^*), t, \epsilon^*, y} \Big[
    \big\| \epsilon^* - \epsilon_\theta^*(z_t^*, t, \tau(y)) \big\|_2^2
\Big], \tag{3}
\end{equation}

\begin{equation}
\mathcal{L}_i = \mathbb{E}_{\mathcal{E}(x_i), t, \epsilon_i, y} \Big[
    \big\| \epsilon_i - \epsilon_\theta^i(z_t^i, t, \tau(y)) \big\|_2^2
\Big].  \tag{4}
\end{equation}
In Stage 3, we jointly train the panoramic alignment module using both the multi-view module and the panoramic module. The corresponding loss is a weighted sum of the two aforementioned losses, formulated as follows.
\begin{equation}
\mathcal{L} = \lambda\mathcal{L}^* + \frac{1}{N} \sum_{i=0}^{N-1} \mathcal{L}_i. \tag{5}
\end{equation} 
$\lambda$ is the weight term to balance the two losses, and $N$ is the number of views.

\begin{table*}[t]
    \centering
    \small
    \caption{The statistic comparison: To our knowledge, Street360 is among the first street scene generation dataset with high-resolution images and additional structural information.}
    \vspace{-0.2cm}
    \begin{tabular}{l|c|c|c|c|c}
    \toprule
        Dataset & Location & Scene type & Image number & Resolution & Condition \\
    \midrule
        CVRG-Pano      & outdoor & countryside road & 600    & 2K    & text \\
        Matterport3D   & indoor  & house            & 10,800 & 2K    & text \\
        Scannet        & indoor  & house            & 1,613  & 2K    & text \\
        HDR360-UHD     & indoor, outdoor & house, wild & 4,394  & 4K--8K & text \\
        Street360      & outdoor & urban street     & 10,000 & 4K--8K & text, seg., contour \\
    \bottomrule
    \end{tabular}
    \vspace{-0.3cm}
    \label{tab:dataset_comparison}
\end{table*}

\begin{table*}[t]
\centering
\small
\begin{minipage}{0.49\textwidth}
\centering
\caption{The model architecture ablations.}
\vspace{-0.2cm}
\begin{tabular}{l|cccc}
\toprule
    Method & FID $\downarrow$ & IS $\uparrow$ & CS $\uparrow$ & $\mathrm{OP\_PSNR}$ $\uparrow$ \\
\midrule
    Pano Branch    & 35.24 & 4.51 & 20.20 & 30.98 \\
    Multiview Branch & 19.26 & 6.21 & 24.58 & 27.29 \\
    Both (Ours)    & \textbf{10.96} & \textbf{6.37} & \textbf{24.69} & \textbf{39.12} \\
\bottomrule
\end{tabular}

\label{table:architecture}
\end{minipage}
\hfill
\begin{minipage}{0.48\textwidth}
\centering
\caption{The prompt insertion ablations.}
\vspace{-0.2cm}
\begin{tabular}{l|ccc}
\toprule
    Method & FID $\downarrow$ & IS $\uparrow$ & CS $\uparrow$ \\
\midrule
    Add             & 23.15 & 5.09 & 20.07 \\
    Concatenation   & 22.51 & 5.42 & 23.76 \\
    ControlNet+Attn (Ours) & \textbf{10.96} & \textbf{6.37} & \textbf{24.69} \\
\bottomrule
\end{tabular}
\label{table:prompt}
\end{minipage}
\end{table*}

\subsection{Structural Information Fusion} 
We adopt three methods to integrate structural information into our Panorama--Perspective Synergy Framework. The first method involves encoding the structural information using an encoder to obtain its latent space features, which are then directly added to the latent space noise, denoted as ``Add''. The second method involves adding viewpoint structural information to the channels during the viewpoint noise input in the multi-view module, and similarly, adding panoramic structural information to the channels during the panoramic noise input in the panoramic module, denoted as ``Concatenation''. The third method involves inputting the viewpoint and panoramic structural information separately into ControlNet \cite{zhang2023controlnet}. We apply cross-attention between the features and the UNet in the corresponding module, denoted as ``ControlNet+Attn''. ControlNet+Attn achieves the best results because it effectively integrates control mechanisms with attention, enabling more precise feature alignment and improved representation, as shown in the prompt insertion ablation study.\par


\subsection{Consistency-Reward Fine-Tuning}
\label{subsec:grpo}
Stages 1--3 optimize per-view denoising losses (Eqs.~(3)--(5)) without directly penalizing disagreement among the final views. After completing the three-stage training procedure, we further apply consistency-reward fine-tuning (CRFT) to the final StreetDiff model using GRPO\cite{shao2024grpo,liu2025flowgrpo}. Given a prompt, we sample a group of $G$ multi-view outputs and optimize
\begin{equation}
\mathcal{L}_{\mathrm{GRPO}}
=
-\mathbb{E}\!\left[
\min\!\left(
\rho_t A,\,
\operatorname{clip}(\rho_t,1-\epsilon,1+\epsilon)A
\right)\right]
+\beta\mathcal{R}_{\mathrm{KL}},
\tag{6}
\end{equation}
where $\rho_t$ is the importance ratio, $A$ is the group-normalized advantage, and $\mathcal{R}_{\mathrm{KL}}$ regularizes the policy toward the frozen Stage-3 model. We define the reward as
$R=\lambda_{\mathrm{geo}}R_{\mathrm{geo}}+\lambda_{\mathrm{sem}}R_{\mathrm{sem}}$, where $R_{\mathrm{geo}}$ measures agreement between the geometrically aligned $45^\circ$ overlapping regions of adjacent views, and $R_{\mathrm{sem}}$ is the CLIP score for preserving prompt alignment. Thus, PAM determines \textit{where} cross-view information is aligned, whereas CRFT directly optimizes \textit{how consistent} the generated views are.

\section{Experiments and Results}
\label{sec:experiments}

\subsection{Dataset Generation and Experiment Setting}

\textbf{Dataset Generation.}
Currently, there are only a few panoramic and multi-view generation datasets specifically designed for indoor tasks, such as ScanNet \cite{dai2017scannet} and Matterport3D \cite{Matterport3D}. However, outdoor datasets, especially street scene datasets, remain largely unavailable. \cite{Learning_to_predict_indoor_illumination_from_a_single_image} contains only 2,100 indoor scenes, while \cite{Zhang2017LearningHD} offered approximately 200 outdoor HDR panoramas. HDR360-UHD \cite{chen2022text2light} comprises 1893 outdoor images and 2501 indoor images, while SUN360 \cite{xiao2012sun360}, similar to HDR360-UHD, is also primarily composed of indoor scenes with relatively few outdoor images. To develop a high-quality street scene generation model, we collected a dataset of 10,000 panoramic/multi-view images covering dozens of regions from the web and combined it with the aforementioned datasets. This resulted in a new high-quality HDR panoramic/multi-view dataset, Street360, containing 10,000 HDR panoramas with resolutions ranging from 4096 $\times$ 2048 (4K) to 8192 $\times$ 4096 (8K). For a more comprehensive dataset comparison, please refer to Table \ref{tab:dataset_comparison}.

\par
Specifically, most images in our dataset were collected from panoramic resources available on the web, with some examples obtained from public platforms such as Google Maps. We then performed view splitting and conditional generation on these panoramas. In particular, we set the view-splitting parameters following the strategy of MVDiffusion: each panorama was divided into six skybox images, and from the four non-polar views, we further extracted eight perspective images. Based on these splits, we employed OneFormer \cite{jain2023oneformer} to generate corresponding panoramic segmentation maps, applied Canny edge detection \cite{canny1986computational} to obtain contour maps, and used BLIP-3 \cite{xue2024xgenmmblip3familyopen} to produce textual descriptions as conditional inputs, thereby providing diverse supervision signals for subsequent multimodal training.

\textbf{Comparison methods.}
We compare our StreetDiff with the following comparison methods, including multi-view and panorama generation SOTAs:
{MVDiffusion} \cite{MVDiffusion}, {PanFusion} \cite{panfusion2024}, {SD+LoRA} \cite{hu2022lora,Rombach_2022_CVPR}, {Text2Light} \cite{chen2022text2light,Rombach_2022_CVPR}.
In the experiments, we compare with these methods by adding the same structure prompts as our method.


\textbf{Implementation details.}
For text-conditioned generation, we adopt the same training and test schedules as MVDiffusion and PanFusion. For generations conditioned on segmentation and contour maps, we additionally train a ControlNet using the same prompt fusion strategy to ensure fair comparison for all methods.

\par
\textbf{Evaluation metrics.}  
We evaluate the generated results using both automatic metrics and a user study. For \textbf{image quality}, we follow prior work and report FID\cite{FID}, IS, and CLIP Score. To assess \textbf{multi-view consistency}, we adopt overlapping PSNR from MVDiffusion. Finally, we conduct a \textbf{user study} where participants compare panoramas generated by different methods. 

\begin{table}[t]
\small
\centering
\caption{The result comparison on the Street360 dataset with different prompts, where `seg' and `cont' refer to using segmentation maps and contour maps as the structure prompts, respectively. \textbf{Bold} refers to the best and \underline{underline} indicates the second best.
}
\vspace{0.2cm}
\setlength{\tabcolsep}{3pt}
\resizebox{\columnwidth}{!}{%
\begin{tabular}{l|c|cccc}
\toprule
    Method      & Prompts         & FID $\downarrow$      & IS $\uparrow$      & CS $\uparrow$  & $\mathrm{OP\_PSNR}$ $\uparrow$  \\

\midrule
     SD+LoRA\_seg       & Text+Seg     & 33.78 & 4.83 & 22.48 & -\\
     PanFusion\_seg       & Text+Seg         &  \underline{27.19} &  \underline{4.92} & 22.10 &-\\
     MVDiffusion\_seg     & Text+Seg         & 29.18 & 4.78 & \underline{23.12} &\underline{35.66}\\
     StreetDiff\_seg (Ours)          & Text+Seg           & \textbf{10.96} & \textbf{6.37} & \textbf{ 24.69}& \textbf{ 39.56 } \\

\midrule
    SD+LoRA\_cont      & Text+Contour  & 35.41 & 4.75 & 20.12 & -\\
    PanFusion\_cont    & Text+Contour  & \underline{28.30} & 3.78 & 17.18 & -\\
    MVDiffusion\_cont  & Text+Contour  & 31.72 & \underline{5.03} & \underline{25.12} & \underline{36.21}\\
    StreetDiff\_cont (Ours) & Text+Contour & \textbf{12.09} & \textbf{5.92} & \textbf{25.21} & \textbf{39.79}\\
\bottomrule
\end{tabular}
}
\vspace{-0.2cm}
\label{table:main_results}
\end{table}

\subsection{Street Generation Results}

\textbf{Quantitative Results.}
Table \ref{table:main_results} presents the quantitative results of using different prompts. 
Using \textbf{segmentation maps} as structure prompts, we significantly outperform SD+LoRA, PanFusion, and MVDiffusion on FID. Besides, we are the best on the IS and CS metrics, indicating that our model excels in generating diverse objects. In contrast, MVDiffusion tends to avoid generating unexpected objects, often resulting in a large number of repeated items. This repetition may enhance alignment with textual prompts, leading to a slightly higher CS score than SD+LoRA and PanFusion, but still lower than ours. Note that SD+LoRA and PanFusion directly generate full panoramic images under this prompt, so their overlapping regions are identical after cropping and $\mathrm{OP\_PSNR}$ is inapplicable (marked ``--'' in Table~\ref{table:main_results}).
Using \textbf{contour maps} as structure prompts, our method also achieves the best results among all methods, indicating that our method could perform robustly under different structure prompts. Considering that segmentation maps provide more structural constraints than contour maps, our method with contour maps performs slightly worse than with segmentation maps, but still outperforms others.

\par \textbf{Qualitative Results.}  
Fig. ~\ref{fig:sem_genertaion} show results with \textbf{segmentation maps} as structure prompts, highlighting three aspects. \textit{Stylistic similarity}: both MVDiffusion and our method generate outputs close to the ground truth. \textit{Image plausibility}: our framework captures local details and global context, avoiding distorted lines, unnatural objects, and panoramic inconsistencies. \textit{Multi-view consistency}: by integrating cues across views, our method achieves smooth transitions and coherent semantics. Overall, it produces the most realistic panoramas with minimal structural distortions (see Fig.~\ref{fig:sem_genertaion}).  


\begin{table}[t]
\small
\centering
\setlength{\tabcolsep}{6pt} %
\caption{User study on 3 metrics: style consistency, realism, and multi-view consistency.}
\vspace{-0.2cm}
\label{tab:userstudy}
\begin{tabular}{lccc}
\hline
\textbf{Method} & \textbf{Style Consist.} & \textbf{Realism} & \textbf{Multi-view Consist.} \\
\hline
MVDiffusion & 16.69\% & 19.84\% & 18.82\% \\
PanFusion   & 1.78\%  & 2.45\%  & 2.88\%  \\
Ours        & \textbf{81.53\%} & \textbf{77.71\%} & \textbf{78.3\%} \\
\hline
\end{tabular}
\vspace{-0.6cm}
\end{table}



\textbf{User Study Results.}
We conduct a user study to compare the performance of all methods using segmentation maps as structure prompts. We use the same prompt to generate \textcolor{black}{45} sets of panoramic images.
Participants are asked to select the best image based on three criteria: Style Consistency, Reasonableness, and Multi-view Consistency. A total of \textcolor{black}{100} valid questionnaires were collected, and Table \ref{tab:userstudy} illustrates that over \textcolor{black}{77}\% of the users regard our method as generating the best street scene images according to all three criteria.
This also proves the advantage of the proposed StreetDiff model in generating more consistent urban street scenes with more realistic object structures.

\subsection{Ablation Study}


\textbf{Model architecture.}
We compare the results of the standalone panoramic branch, the standalone multi-view branch, and our full Panorama--Perspective Synergy Framework in Table \ref{table:architecture}.
It is observed that the panoramic branch exhibits a worse FID score, indicating that the authenticity of the generated images is insufficient. This is primarily because Stable Diffusion (SD) performs optimally at lower resolutions (e.g., 512 $\times$ 512), and its performance naturally declines at higher resolutions. However, since the panoramic branch performs denoising on the same noise, the generated images display good overall consistency, suggesting that the panoramic branch possesses certain global consistency features, though it still lacks structural consistency. In contrast, the multi-view branch achieves a better FID score than the panoramic branch, as the image resolution generated by the multi-view branch aligns with SD's strengths. Nevertheless, the consistency between adjacent views is relatively poor due to the simultaneous denoising of multiple noises, leading to suboptimal adjacent view images. Overall, our method not only achieves optimal performance in terms of generation quality but also attains the best consistency results. This proves our model fully leverages SD's powerful generation capabilities, captures global consistency features in the panoramic branch, and effectively controls structural information, thereby generating realistic images that conform to spatial structures and high consistency.

\begin{table}[t]
\small
\caption{The results on the Street360 dataset with only text prompts.}
\centering
\setlength{\tabcolsep}{3pt}
\begin{tabular}{lcccc}
\toprule
    Method & FID $\downarrow$ & IS $\uparrow$ & CS $\uparrow$ & $\mathrm{OP\_PSNR}$ $\uparrow$ \\
\midrule
    SD+LoRA       & 30.19 & 5.06 & 20.46 & -\\
    Text2Light    & 113.09 & 5.06 & 21.72 & -\\
    PanFusion     & 18.55 & 4.89 & 21.48 & -\\
    MVDiffusion   & 24.59 & \underline{5.83} & \underline{24.10} & 38.65\\
    CubeDiff
& \underline{16.82}
& 5.71
& 23.86
& \underline{38.91} \\

    \midrule
    StreetDiff (Ours) & \textbf{11.95} & \textbf{6.02} & \textbf{24.93} & \textbf{39.12}\\
\bottomrule
\end{tabular}
\label{table:text_prompts}
\end{table}



\textbf{Prompt insertion.}
We conduct an ablation study on different ways of incorporating structural information (Table \ref{table:prompt}). A simple approach concatenates structural information with the noise input as an additional channel. Another approach adopts ControlNet, where each layer output is added to the corresponding layers in the U-Net. Our method instead integrates ControlNet features into the U-Net via cross-attention, which achieves the best results. This demonstrates that cross-attention effectively leverages structural cues, enhancing spatial perception and leading to images that better match the expected outcomes.


\textbf{Only using text prompts.}
We compare all methods using only text prompts in Table~\ref{table:text_prompts}. Since the official implementation of CubeDiff is unavailable, its results are obtained using the publicly available unofficial OpenCubeDiff implementation. Even without structure prompts, StreetDiff achieves the best overall performance on Street360, demonstrating that panoramic features effectively guide multi-view generation and improve both cross-view consistency and visual details.


\textbf{Ablation study on PAM.}
To verify the effectiveness of PAM, we compare it with two simplified cross-branch interaction designs (Table \ref{tab:pam_ablation}): \textbf{(i) Naive Cross-Attention}, which directly applies standard cross-attention from the multi-view branch to panoramic features to absorb global context without any projection-aware constraint; and \textbf{(ii) Sequential Interaction}, which first uses the multi-view branch to refine the panoramic branch and then uses the refined panoramic features to guide the multi-view branch. Both variants remove the projection-consistent point-to-region alignment in PAM and serve to isolate the contribution of geometry-aware correspondence modeling.


\begin{figure*}[t]
    \centering
    \includegraphics[width=0.8\textwidth]{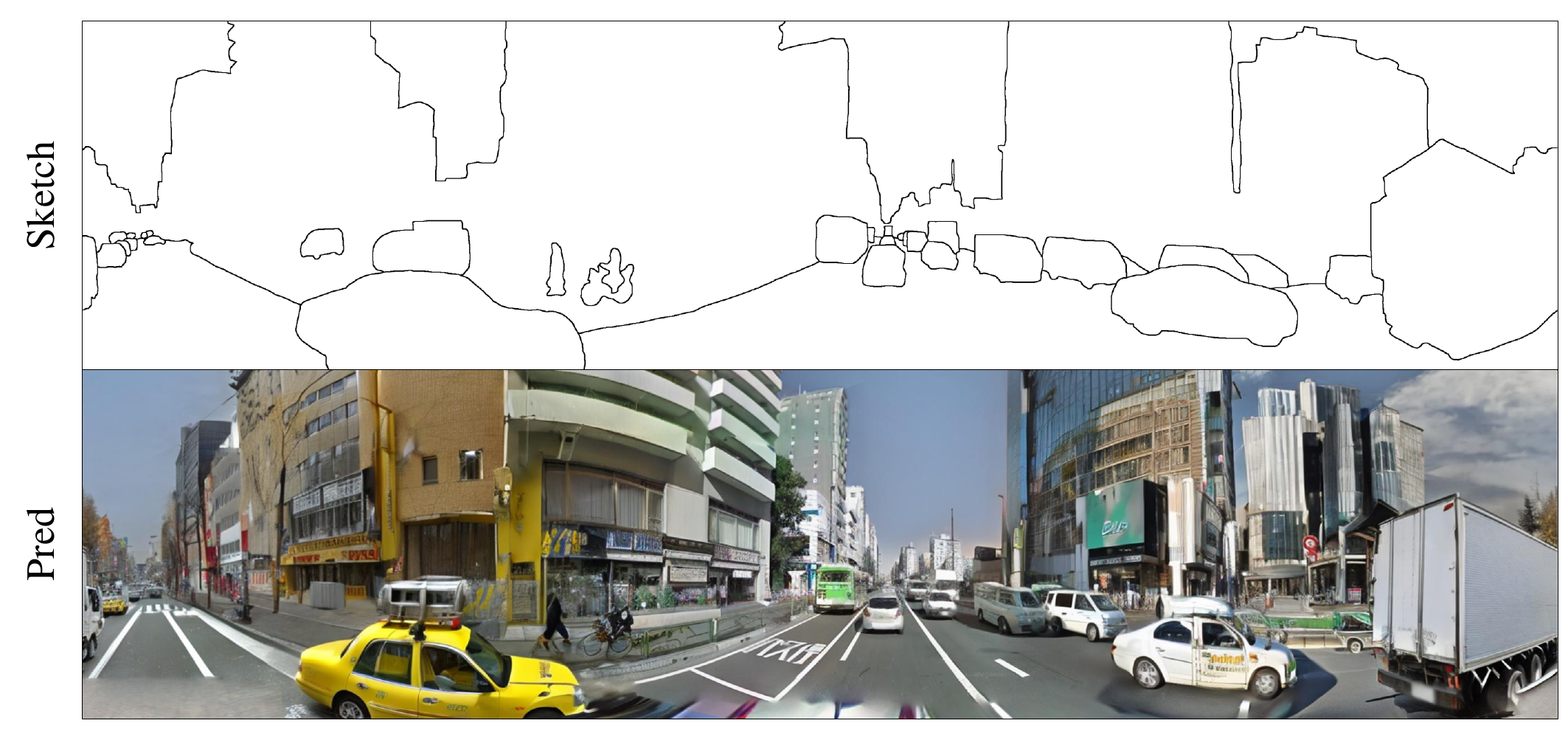}
    \vspace{-0.25cm}
    \caption{Street generation with user sketches: The model is trained with contour maps as structure prompts, while user sketches are used only at test time.}
    \label{fig:sketch}
    \vspace{-0.4cm}
\end{figure*}

\textbf{Training Strategy Ablation.}
To evaluate the effectiveness of our staged training strategy, we conduct an ablation comparing two training configurations: (i) end-to-end training, where the entire dual-branch architecture is optimized jointly from scratch, and (ii) our 3-stage training strategy, which progressively learns panoramic priors, strengthens intra-view consistency in the multi-view branch, and finally aligns both branches jointly.
As in Table \ref{tab:traing_strategy}, the end-to-end variant suffers from significant performance degradation in both visual quality and cross-view consistency. In contrast, our staged strategy achieves noticeably better convergence behavior and higher overall fidelity. These results validate that the progressive learning scheme is not merely an implementation convenience, but a crucial component enabling stable and consistency-aware generation.

\begin{table}[t]
    \centering
    \caption{Ablation of PAM variants on Street360 with text-only prompts.}
    \label{tab:pam_ablation}
    \scriptsize
    \setlength{\tabcolsep}{2pt}
    \resizebox{\columnwidth}{!}{%
    \begin{tabular}{l|cccc}
        \toprule
        Method & FID$\downarrow$ & IS$\uparrow$ & CS$\uparrow$ & $\mathrm{OP\_PSNR}\uparrow$ \\
    \midrule
        Naive (MV$\rightarrow$Pano) & 19.21 & 5.02 & 22.54 & 37.28 \\
        Sequential (MV$\rightarrow$Pano$\rightarrow$MV) & 21.37 & 5.36 & 23.71 & 36.21 \\
        PAM (Ours) & \textbf{11.95} & \textbf{6.02} & \textbf{24.93} & \textbf{39.12} \\
    \bottomrule
    \end{tabular}%
    }
\end{table}

\begin{table}[t]
    \centering
    \caption{The training strategy ablation study. The experiments are conducted on the Street360 dataset with only text prompts.}
    \label{tab:traing_strategy}
    \small
    \setlength{\tabcolsep}{3pt}
    \begin{tabular}{l|cccc}
        \toprule
        Method & FID $\downarrow$ & IS $\uparrow$ & CS $\uparrow$ & $\mathrm{OP\_PSNR}$ $\uparrow$ \\
        \midrule
        end-to-end & 22.35 & 4.98 & 22.25 & 38.71 \\
        3-stage (Ours) & \textbf{11.95} & \textbf{6.02} & \textbf{24.93} & \textbf{39.12} \\
        \bottomrule
    \end{tabular}
\end{table}

\textbf{Consistency-reward fine-tuning.}
Table \ref{tab:rl} evaluates the stage of Sec.~\ref{subsec:grpo} under the text-only setting, so the first row is our Stage-3 model in Table \ref{table:text_prompts}. GRPO performs best on all metrics, while DDPO \cite{black2024ddpo} and Diffusion-DPO \cite{wallace2024diffusiondpo} yield smaller gains in cross-view consistency; rewarding semantics alone raises CS but leaves $\mathrm{OP\_PSNR}$ essentially unchanged, indicating that consistency has to be optimized explicitly rather than obtained as a by-product of better text alignment. Removing the KL term yields a higher reward and even a higher $\mathrm{OP\_PSNR}$, but sharply degrades FID, as the model converges to over-smoothed facades that trivially maximize overlap agreement, a clear case of reward hacking; we therefore keep a moderate $\beta$.

\begin{table}[t]
\centering
\caption{Consistency-reward fine-tuning on Street360 with text-only prompts.}
\label{tab:rl}
\small
\setlength{\tabcolsep}{3pt}
\begin{tabular}{l|cccc}
\toprule
    Method & FID$\downarrow$ & IS$\uparrow$ & CS$\uparrow$ & $\mathrm{OP\_PSNR}\uparrow$ \\
\midrule
    StreetDiff (w/o RL)                               & 11.95 & 6.02 & 24.93 & 39.12 \\
    \; + DDPO \cite{black2024ddpo}                    & 12.86 & 5.88 & 24.71 & 39.94 \\
    \; + Diffusion-DPO \cite{wallace2024diffusiondpo} & 11.62 & 6.11 & 25.18 & 40.37 \\
    \; + GRPO (Ours)                                  & \textbf{11.08} & \textbf{6.29} & \textbf{25.64} & \textbf{41.23} \\
\midrule
    \; GRPO, $R_{\text{sem}}$ only                    & 11.79 & 6.05 & 25.71 & 39.28 \\
    \; GRPO, w/o KL ($\beta{=}0$)                     & 15.73 & 5.74 & 26.12 & 42.05 \\
\bottomrule
\end{tabular}
\end{table}

\textbf{Street generation with user sketches.} We use user-input sketches as structure prompts in Fig.~\ref{fig:sketch}. The model is trained with contour maps as structure prompts, while user sketches are used only at test time. The results show that StreetDiff can generate high-quality images that follow the sketch layout, suggesting potential use in interactive street scene creation.

\section{Discussion and Conclusion}
\label{sec:conclusion}

In this paper, we propose an urban street scene generation model StreetDiff, which can generate highly consistent and realistic multi-view street scene images. The model employs structure prompts and a Panorama--Perspective Synergy Framework to make the Stable Diffusion model deal with the object structure challenges well in street scene generation. We also propose a large multi-view street scene dataset Street360 for studying the issue.
The experiments demonstrate that the proposed StreetDiff model has a clear advantage on the street scene generation task over existing multi-view generation models designed for indoor or wild applications. 
Besides, in addition to segmentation maps and contour maps,  our model can also accept user-input sketches and generate high-quality images aligning with the user input.
The proposed model and dataset shall advance the multi-view generation tasks to more complicated scenarios. 
In future work, we plan to explore more fine-grained structural guidance, such as user-defined semantic bounding boxes and bird's-eye view (BEV) representations, to broaden the applicability of urban street scene generation.

\section*{Acknowledgements}
This work was supported by National Key R\&D Program of China (2024YFB3908500, 2024YFB3908504), NSFC (62202312), ICFCRT (W2441020), Shenzhen Science and Technology Program (KJZD20240903100022028, KQTD20210811090044003), Scientific Development Fund from Guangdong Provincial Key Laboratory of Visual Media and Multidimensional Intelligence, and Scientific Foundation for Youth Scholars of Shenzhen University.

\appendices

\bibliographystyle{IEEEtran}
\bibliography{streetdiff}

@String(CVPR= {IEEE Conf. Comput. Vis. Pattern Recog.})

@String(ICCV= {Int. Conf. Comput. Vis.})

@String(ECCV= {Eur. Conf. Comput. Vis.})

@String(NIPS= {Adv. Neural Inform. Process. Syst.})

@String(TOG= {ACM Trans. Graph.})

@String(ICLR = {Int. Conf. Learn. Represent.})

@String(VR   = {Vis. Res.})

@String(CVPR  = {CVPR})

@String(ICCV  = {ICCV})

@String(ECCV  = {ECCV})

@String(NIPS  = {NeurIPS})

@String(TOG   = {ACM TOG})

@String(ICLR  = {ICLR})

@article{shao2024grpo,
  title={{DeepSeekMath}: Pushing the limits of mathematical reasoning in open language models},
  author={Shao, Zhihong and Wang, Peiyi and Zhu, Qihao and Xu, Runxin and Song, Junxiao and Bi, Xiao and Zhang, Haowei and Zhang, Mingchuan and Li, Y K and Wu, Y and Guo, Daya},
  journal={arXiv preprint arXiv:2402.03300}, year={2024}
}

@inproceedings{liu2025flowgrpo,
 author = {Liu, Jie and Liu, Gongye and Liang, Jiajun and Li, Yangguang and Liu, Jiaheng and Wang, Xintao and Wan, Pengfei and ZHANG, Di and Ouyang, Wanli},
 booktitle = {Advances in Neural Information Processing Systems},
 doi = {10.52202/085713-1362},
 editor = {D. Belgrave and C. Zhang and H. Lin and R. Pascanu and P. Koniusz and M. Ghassemi and N. Chen},
 pages = {40783--40818},
 publisher = {Curran Associates, Inc.},
 title = {Flow-GRPO: Training Flow Matching Models via Online RL},
 url = {https://proceedings.neurips.cc/paper_files/paper/2025/file/3a10c46572628d58cb44fb705f25cbbf-Paper-Conference.pdf},
 volume = {38, Main Conference},
 year = {2025}
}

@inproceedings{black2024ddpo,
  title={Training diffusion models with reinforcement learning},
  author={Black, Kevin and Janner, Michael and Du, Yilun and Kostrikov, Ilya and Levine, Sergey},
  booktitle={International Conference on Learning Representations (ICLR)}, year={2024}
}

@inproceedings{wallace2024diffusiondpo,
  title={Diffusion model alignment using direct preference optimization},
  author={Wallace, Bram and Dang, Meihua and Rafailov, Rafael and Zhou, Linqi and Lou, Aaron and Purushwalkam, Senthil and Ermon, Stefano and Xiong, Caiming and Joty, Shafiq and Naik, Nikhil},
  booktitle={Proceedings of the IEEE/CVF Conference on Computer Vision and Pattern Recognition (CVPR)}, year={2024}
}

@INPROCEEDINGS{Geometry-Guided-Cross-View-Diffusion,
  author={Lin, Tao Jun and Wang, Wenqing and Shi, Yujiao and Perincherry, Akhil and Vora, Ankit and Li, Hongdong},
  booktitle={2025 International Conference on 3D Vision (3DV)}, 
  title={Geometry-Guided Cross-View Diffusion for One-to-Many Cross-View Image Synthesis}, 
  year={2025},
  volume={},
  number={},
  pages={866-881},
  doi={10.1109/3DV66043.2025.00085}}

@inproceedings{yan2024streetcrafter,
  title={StreetCrafter: Street View Synthesis with Controllable Video Diffusion Models},
  author={Yan, Yunzhi and Xu, Zhen and Lin, Haotong and Jin, Haian and Guo, Haoyu and Wang, Yida and Zhan, Kun and Lang, Xianpeng and Bao, Hujun and Zhou, Xiaowei and Peng, Sida},
  booktitle={Proceedings of the IEEE/CVF Conference on Computer Vision and Pattern Recognition (CVPR)},
  year={2025},
}

@article{ze2025controllable,
  title={Controllable Satellite-to-Street-View Synthesis with Precise Pose Alignment and Zero-Shot Environmental Control},
  author={Ze, Xianghui and Song, Zhenbo and Wang, Qiwei and Lu, Jianfeng and Shi, Yujiao},
  journal={arXiv preprint arXiv:2502.03498},
  year={2025}
}

@InProceedings{ye2024skydiffusion,
    author    = {Ye, Junyan and He, Jun and Li, Weijia and Lv, Zhutao and Lin, Yi and Yu, Jinhua and Yang, Haote and He, Conghui},
    title     = {Leveraging BEV Paradigm for Ground-to-Aerial Image Synthesis},
    booktitle = {Proceedings of the IEEE/CVF International Conference on Computer Vision (ICCV)},
    month     = {October},
    year      = {2025},
    pages     = {28451-28461}
}

@misc{li2024crossviewdiffcrossviewdiffusionmodel,
      title={CrossViewDiff: A Cross-View Diffusion Model for Satellite-to-Street View Synthesis}, 
      author={Weijia Li and Jun He and Junyan Ye and Huaping Zhong and Zhimeng Zheng and Zilong Huang and Dahua Lin and Conghui He},
      year={2024},
      eprint={2408.14765},
      archivePrefix={arXiv},
      primaryClass={cs.CV},
      url={https://arxiv.org/abs/2408.14765}, 
}

@InProceedings{zheng2025panorama,
      title={Panorama Generation From NFoV Image Done Right},
      author={Zheng, Dian and Zhang, Cheng and Wu, Xiao-Ming and Li, Cao and Lv, Chengfei and Hu, Jian-Fang and Zheng, Wei-Shi},
      booktitle={Proceedings of the IEEE/CVF Conference on Computer Vision and Pattern Recognition},
      year={2025}
}

@inproceedings{yang2024layerpano3d,
author = {Yang, Shuai and Tan, Jing and Zhang, Mengchen and Wu, Tong and Wetzstein, Gordon and Liu, Ziwei and Lin, Dahua},
title = {LayerPano3D: Layered 3D Panorama for Hyper-Immersive Scene Generation},
year = {2025},
isbn = {9798400715402},
publisher = {Association for Computing Machinery},
address = {New York, NY, USA},
doi = {10.1145/3721238.3730643},
booktitle = {Proceedings of the Special Interest Group on Computer Graphics and Interactive Techniques Conference Conference Papers},
articleno = {109},
numpages = {10},
series = {SIGGRAPH Conference Papers '25}
}

@InProceedings{Sun_2025_CVPR,
    author    = {Sun, Xiancheng and Xu, Mai and Li, Shengxi and Ma, Senmao and Deng, Xin and Jiang, Lai and Shen, Gang},
    title     = {Spherical Manifold Guided Diffusion Model for Panoramic Image Generation},
    booktitle = {Proceedings of the Computer Vision and Pattern Recognition Conference (CVPR)},
    month     = {June},
    year      = {2025},
    pages     = {5824-5834}
}

@article{xiong2024mixed,
  title     = {{Mixed-View Panorama Synthesis Using Geospatially Guided Diffusion}},
  author    = {Xiong, Zhexiao and Xing, Xin and Workman, Scott and Khanal, Subash and Jacobs, Nathan},
  journal   = {Transactions on Machine Learning Research},
  year      = {2025},
  url       = {https://mlanthology.org/tmlr/2025/xiong2025tmlr-mixedview/}
}

@inproceedings{deng2024streetscapes,
  title     = {Streetscapes: Large-scale Consistent Street View Generation
               Using Autoregressive Video Diffusion},
  author    = {Deng, Boyang and Tucker, Richard and Li, Zhengqi
               and Guibas, Leonidas and Snavely, Noah and Wetzstein, Gordon},
  booktitle = {SIGGRAPH 2024 Conference Papers},
  year      = {2024}
}

@article{gao2024opamatextguidedmamba,
title = {OPa-Ma: Text guided Mamba for 360-degree image out-painting},
journal = {Pattern Recognition},
volume = {180},
pages = {113957},
year = {2026},
issn = {0031-3203},
doi = {https://doi.org/10.1016/j.patcog.2026.113957},
url = {https://www.sciencedirect.com/science/article/pii/S0031320326009222},
author = {Penglei Gao and Kai Yao and Tiandi Ye and Steven Z. Wang and Yuan Yao and Xiaofeng Wang}
}

@InProceedings{zhang2024perldiff,
    author    = {Zhang, Jinhua and Sheng, Hualian and Cai, Sijia and Deng, Bing and Liang, Qiao and Li, Wen and Fu, Ying and Ye, Jieping and Gu, Shuhang},
    title     = {PerLDiff: Controllable Street View Synthesis Using Perspective-Layout Diffusion Model},
    booktitle = {Proceedings of the IEEE/CVF International Conference on Computer Vision (ICCV)},
    month     = {October},
    year      = {2025},
    pages     = {26306-26315}
}

@inproceedings{zhange2023diffcollage,
    title={DiffCollage: Parallel Generation of Large Content with Diffusion Models},
    author={Qinsheng Zhang and Jiaming Song and Xun Huang and Yongxin Chen and Ming-yu Liu},
    booktitle={CVPR},
    year={2023}
}

@inproceedings{IS,
author = {Salimans, Tim and Goodfellow, Ian and Zaremba, Wojciech and Cheung, Vicki and Radford, Alec and Chen, Xi},
title = {Improved techniques for training GANs},
year = {2016},
isbn = {9781510838819},
publisher = {Curran Associates Inc.},
address = {Red Hook, NY, USA},
pages = {2234--2242},
numpages = {9},
location = {Barcelona, Spain},
series = {NIPS'16}
}

@inproceedings{FID,
author = {Heusel, Martin and Ramsauer, Hubert and Unterthiner, Thomas and Nessler, Bernhard and Hochreiter, Sepp},
title = {GANs trained by a two time-scale update rule converge to a local nash equilibrium},
year = {2017},
isbn = {9781510860964},
publisher = {Curran Associates Inc.},
address = {Red Hook, NY, USA},
pages = {6629--6640},
numpages = {12},
location = {Long Beach, California, USA},
series = {NIPS'17}
}

@article{canny1986computational,
  title={A Computational Approach to Edge Detection},
  author={Canny, John},
  journal={IEEE Transactions on Pattern Analysis and Machine Intelligence},
  volume={8},
  number={6},
  pages={679--698},
  year={1986},
  publisher={IEEE}
}

@InProceedings{xue2024xgenmmblip3familyopen,
    author    = {Xue, Le and Shu, Manli and Awadalla, Anas and Wang, Jun and Yan, An and Purushwalkam, Senthil and Zhou, Honglu and Prabhu, Viraj and Dai, Yutong and Ryoo, Michael S and Kendre, Shrikant and Zhang, Jieyu and Tseng, Shaoyen and Lujan-Moreno, Gustavo Adolfo and Olson, Matthew Lyle and Hinck, Musashi and Cobbley, David and Lal, Vasudev and Qin, Can and Zhang, Shu and Chen, Chia-Chih and Yu, Ning and Tan, Juntao and Awalgaonkar, Tulika Manoj and Heinecke, Shelby and Wang, Huan and Choi, Yejin and Schmidt, Ludwig and Chen, Zeyuan and Savarese, Silvio and Niebles, Juan Carlos and Xiong, Caiming and Xu, Ran},
    title     = {BLIP-3: A Family of Open Large Multimodal Models},
    booktitle = {Proceedings of the IEEE/CVF International Conference on Computer Vision (ICCV) Workshops},
    month     = {October},
    year      = {2025},
    pages     = {6183-6194}
}

@INPROCEEDINGS{xiao2012sun360,
  author={Xiao, Jianxiong and Ehinger, Krista A. and Oliva, Aude and Torralba, Antonio},
  booktitle={2012 IEEE Conference on Computer Vision and Pattern Recognition}, 
  title={Recognizing scene viewpoint using panoramic place representation}, 
  year={2012},
  volume={},
  number={},
  pages={2695-2702},
  doi={10.1109/CVPR.2012.6247991}}

@article{Zhang2017LearningHD,
  title={Learning High Dynamic Range from Outdoor Panoramas},
  author={Jinsong Zhang and Jean-Fran\c{c}ois Lalonde},
  journal={2017 IEEE International Conference on Computer Vision (ICCV)},
  year={2017},
  pages={4529-4538},
  url={https://api.semanticscholar.org/CorpusID:226521}
}

@article{Learning_to_predict_indoor_illumination_from_a_single_image,
author = {Gardner, Marc-Andr\'{e} and Sunkavalli, Kalyan and Yumer, Ersin and Shen, Xiaohui and Gambaretto, Emiliano and Gagn\'{e}, Christian and Lalonde, Jean-Fran\c{c}ois},
title = {Learning to predict indoor illumination from a single image},
year = {2017},
issue_date = {December 2017},
publisher = {Association for Computing Machinery},
address = {New York, NY, USA},
volume = {36},
number = {6},
issn = {0730-0301},
url = {https://doi.org/10.1145/3130800.3130891},
doi = {10.1145/3130800.3130891},
journal = {ACM Trans. Graph.},
month = nov,
articleno = {176},
numpages = {14}
}

@inproceedings{Matterport3D,
  title={Matterport3D: Learning from RGB-D Data in Indoor Environments},
  author={Chang, Angel and Dai, Angela and Funkhouser, Thomas and Halber, Maciej and Niessner, Matthias and Savva, Manolis and Song, Shuran and Zeng, Andy and Zhang, Yinda},
  booktitle={International Conference on 3D Vision (3DV)},
  year={2017}
}

@inproceedings{dai2017scannet,
  title={ScanNet: Richly-annotated 3D reconstructions of indoor scenes},
  author={Dai, Angela and Chang, Angel X and Savva, Manolis and Halber, Maciej and Funkhouser, Thomas and Nie{\ss}ner, Matthias},
  booktitle={Proceedings of the IEEE conference on computer vision and pattern recognition},
  pages={5828--5839},
  year={2017}
}

@inproceedings{hu2022lora,
title={Lo{RA}: Low-Rank Adaptation of Large Language Models},
author={Edward J Hu and Yelong Shen and Phillip Wallis and Zeyuan Allen-Zhu and Yuanzhi Li and Shean Wang and Lu Wang and Weizhu Chen},
booktitle={International Conference on Learning Representations},
year={2022},
url={https://openreview.net/forum?id=nZeVKeeFYf9}
}

@misc{Equirec2Perspec,
  author = {Fuen Wang},
  title = {Equirec2Perspec: A tool to project equirectangular panorama into perspective images},
  year = {2025},
  howpublished = {\url{https://github.com/fuenwang/Equirec2Perspec}},
  note = {Accessed: 2025-03-05}
}

@InProceedings{PanoFree,
author="Liu, Aoming
and Li, Zhong
and Chen, Zhang
and Li, Nannan
and Xu, Yi
and Plummer, Bryan A.",
editor="Leonardis, Ale{\v{s}}
and Ricci, Elisa
and Roth, Stefan
and Russakovsky, Olga
and Sattler, Torsten
and Varol, G{\"u}l",
title="PanoFree: Tuning-Free Holistic Multi-view Image Generation with Cross-View Self-guidance",
booktitle="Computer Vision -- ECCV 2024",
year="2025",
publisher="Springer Nature Switzerland",
address="Cham",
pages="146--164",
isbn="978-3-031-73383-3"
}

@inproceedings{jain2023oneformer,
      title={{OneFormer: One Transformer to Rule Universal Image Segmentation}},
      author={Jitesh Jain and Jiachen Li and MangTik Chiu and Ali Hassani and Nikita Orlov and Humphrey Shi},
        booktitle="CVPR",
      journal={CVPR}, 
      year={2023}
    }

@InProceedings{Customizing_360-Degree_Panorama_Diffusion_Models,
    author    = {Wang, Hai and Xiang, Xiaoyu and Fan, Yuchen and Xue, Jing-Hao},
    title     = {Customizing 360-Degree Panoramas Through Text-to-Image Diffusion Models},
    booktitle = {Proceedings of the IEEE/CVF Winter Conference on Applications of Computer Vision (WACV)},
    month     = {January},
    year      = {2024},
    pages     = {4933-4943}
}

@inproceedings{360-Degree_Panorama_Generation_from_Few_Unregistered_NFoV_Images,
author = {Wang, Jionghao and Chen, Ziyu and Ling, Jun and Xie, Rong and Song, Li},
title = {360-Degree Panorama Generation from Few Unregistered NFoV Images},
year = {2023},
isbn = {9798400701085},
publisher = {Association for Computing Machinery},
address = {New York, NY, USA},
url = {https://doi.org/10.1145/3581783.3612508},
doi = {10.1145/3581783.3612508},
booktitle = {Proceedings of the 31st ACM International Conference on Multimedia},
pages = {6811--6821},
numpages = {11},
location = {Ottawa ON, Canada},
series = {MM '23}
}

@InProceedings{BIPS,
author="Oh, Changgyoon
and Cho, Wonjune
and Chae, Yujeong
and Park, Daehee
and Wang, Lin
and Yoon, Kuk-Jin",
editor="Avidan, Shai
and Brostow, Gabriel
and Ciss{\'e}, Moustapha
and Farinella, Giovanni Maria
and Hassner, Tal",
title="BIPS: Bi-modal Indoor Panorama Synthesis via Residual Depth-Aided Adversarial Learning",
booktitle="Computer Vision -- ECCV 2022",
year="2022",
publisher="Springer Nature Switzerland",
address="Cham",
pages="352--371",
isbn="978-3-031-19787-1"
}

@inproceedings{StyleLight,
author = {Wang, Guangcong and Yang, Yinuo and Loy, Chen Change and Liu, Ziwei},
title = {StyleLight: HDR Panorama Generation for Lighting Estimation and Editing},
year = {2022},
isbn = {978-3-031-19783-3},
publisher = {Springer-Verlag},
address = {Berlin, Heidelberg},
url = {https://doi.org/10.1007/978-3-031-19784-0_28},
doi = {10.1007/978-3-031-19784-0_28},
booktitle = {Computer Vision -- ECCV 2022: 17th European Conference, Tel Aviv, Israel, October 23--27, 2022, Proceedings, Part XV},
pages = {477--492},
numpages = {16},
location = {Tel Aviv, Israel}
}

@INPROCEEDINGS{GAN360,
  author={Dastjerdi, Mohammad Reza Karimi and Hold-Geoffroy, Yannick and Eisenmann, Jonathan and Khodadadeh, Siavash and Lalonde, Jean-Fran\c{c}ois},
  booktitle={2022 International Conference on 3D Vision (3DV)}, 
  title={Guided Co-Modulated GAN for 360$^\circ$ Field of View Extrapolation}, 
  year={2022},
  volume={},
  number={},
  pages={475-485},
  doi={10.1109/3DV57658.2022.00059}}

@InProceedings{360-Degree_Outpainting,
    author    = {Akimoto, Naofumi and Matsuo, Yuhi and Aoki, Yoshimitsu},
    title     = {Diverse Plausible 360-Degree Image Outpainting for Efficient 3DCG Background Creation},
    booktitle = {Proceedings of the IEEE/CVF Conference on Computer Vision and Pattern Recognition (CVPR)},
    month     = {June},
    year      = {2022},
    pages     = {11441-11450}
}

@InProceedings{latent_nerf,
    author    = {Metzer, Gal and Richardson, Elad and Patashnik, Or and Giryes, Raja and Cohen-Or, Daniel},
    title     = {Latent-NeRF for Shape-Guided Generation of 3D Shapes and Textures},
    booktitle = {Proceedings of the IEEE/CVF Conference on Computer Vision and Pattern Recognition (CVPR)},
    month     = {June},
    year      = {2023},
    pages     = {12663-12673}
}

@article{liu2023one2345,
  title={One-2-3-45: Any single image to 3d mesh in 45 seconds without per-shape optimization},
  author={Liu, Minghua and Xu, Chao and Jin, Haian and Chen, Linghao and Varma T, Mukund and Xu, Zexiang and Su, Hao},
  journal={Advances in Neural Information Processing Systems},
  volume={36},
  year={2024}
}

@InProceedings{tang2024mvdiffusionpp,
author="Tang, Shitao
and Chen, Jiacheng
and Wang, Dilin
and Tang, Chengzhou
and Zhang, Fuyang
and Fan, Yuchen
and Chandra, Vikas
and Furukawa, Yasutaka
and Ranjan, Rakesh",
editor="Leonardis, Ale{\v{s}}
and Ricci, Elisa
and Roth, Stefan
and Russakovsky, Olga
and Sattler, Torsten
and Varol, G{\"u}l",
title="MVDiffusion++: A Dense High-Resolution Multi-view Diffusion Model for Single or Sparse-View 3D Object Reconstruction",
booktitle="Computer Vision -- ECCV 2024",
year="2025",
publisher="Springer Nature Switzerland",
address="Cham",
pages="175--191",
isbn="978-3-031-72640-8"
}

@inproceedings{liu2023zero1to3,
  title={Zero-1-to-3: Zero-shot one image to 3d object},
  author={Liu, Ruoshi and Wu, Rundi and Van Hoorick, Basile and Tokmakov, Pavel and Zakharov, Sergey and Vondrick, Carl},
  booktitle={Proceedings of the IEEE/CVF international conference on computer vision},
  pages={9298--9309},
  year={2023}
}

@misc{zhang2023controlnet,
  title={Adding Conditional Control to Text-to-Image Diffusion Models}, 
  author={Lvmin Zhang and Anyi Rao and Maneesh Agrawala},
  booktitle={IEEE International Conference on Computer Vision (ICCV)},
  year={2023}
}

@inproceedings{DrivingDiffusion,
author = {Li, Xiaofan and Zhang, Yifu and Ye, Xiaoqing},
title={DrivingDiffusion: Layout-Guided multi-view driving scene video generation with latent diffusion model},
year = {2024},
isbn = {978-3-031-73228-7},
publisher = {Springer-Verlag},
address = {Berlin, Heidelberg},
url = {https://doi.org/10.1007/978-3-031-73229-4_27},
doi = {10.1007/978-3-031-73229-4_27},
booktitle = {Computer Vision -- ECCV 2024: 18th European Conference, Milan, Italy, September 29--October 4, 2024, Proceedings, Part LXXVIII},
pages = {469--485},
numpages = {17},
location = {Milan, Italy}
}

@INPROCEEDINGS{yang2023dreamspacedreamingroomspace,
  author={Yang, Bangbang and Dong, Wenqi and Ma, Lin and Hu, Wenbo and Liu, Xiao and Cui, Zhaopeng and Ma, Yuewen},
  booktitle={2024 IEEE Conference Virtual Reality and 3D User Interfaces (VR)}, 
  title={DreamSpace: Dreaming Your Room Space with Text-Driven Panoramic Texture Propagation}, 
  year={2024},
  volume={},
  number={},
  pages={650-660},
  doi={10.1109/VR58804.2024.00085}}

@inproceedings{NIPS_Photorealistic_t2i,
author = {Saharia, Chitwan and Chan, William and Saxena, Saurabh and Lit, Lala and Whang, Jay and Denton, Emily and Ghasemipour, Seyed Kamyar Seyed and Ayan, Burcu Karagol and Mahdavi, S. Sara and Gontijo-Lopes, Raphael and Salimans, Tim and Ho, Jonathan and Fleet, David J and Norouzi, Mohammad},
title = {Photorealistic text-to-image diffusion models with deep language understanding},
year = {2022},
isbn = {9781713871088},
publisher = {Curran Associates Inc.},
address = {Red Hook, NY, USA},
booktitle = {Proceedings of the 36th International Conference on Neural Information Processing Systems},
articleno = {2643},
numpages = {16},
location = {New Orleans, LA, USA},
series = {NIPS '22}
}

@InProceedings{Rombach_2022_CVPR,
    author    = {Rombach, Robin and Blattmann, Andreas and Lorenz, Dominik and Esser, Patrick and Ommer, Bj\"orn},
    title     = {High-Resolution Image Synthesis With Latent Diffusion Models},
    booktitle = {Proceedings of the IEEE/CVF Conference on Computer Vision and Pattern Recognition (CVPR)},
    month     = {June},
    year      = {2022},
    pages     = {10684-10695}
}

@article{ramesh2022hierarchicaltextconditionalimagegeneration,
  title={Hierarchical text-conditional image generation with clip latents},
  author={Ramesh, Aditya and Dhariwal, Prafulla and Nichol, Alex and Chu, Casey and Chen, Mark},
  journal={arXiv preprint arXiv:2204.06125},
  volume={1},
  number={2},
  pages={3},
  year={2022}
}

@misc{nichol2022glidephotorealisticimagegeneration,
      title={GLIDE: Towards Photorealistic Image Generation and Editing with Text-Guided Diffusion Models}, 
      author={Alex Nichol and Prafulla Dhariwal and Aditya Ramesh and Pranav Shyam and Pamela Mishkin and Bob McGrew and Ilya Sutskever and Mark Chen},
      year={2022},
      eprint={2112.10741},
      archivePrefix={arXiv},
      primaryClass={cs.CV},
      journal={arXiv preprint arXiv:2112.10741},
      url={https://arxiv.org/abs/2112.10741}, 
}

@InProceedings{ramesh2021zeroshottexttoimagegeneration,
  title = 	 {Zero-Shot Text-to-Image Generation},
  author =       {Ramesh, Aditya and Pavlov, Mikhail and Goh, Gabriel and Gray, Scott and Voss, Chelsea and Radford, Alec and Chen, Mark and Sutskever, Ilya},
  booktitle = 	 {Proceedings of the 38th International Conference on Machine Learning},
  pages = 	 {8821--8831},
  year = 	 {2021},
  editor = 	 {Meila, Marina and Zhang, Tong},
  volume = 	 {139},
  series = 	 {Proceedings of Machine Learning Research},
  month = 	 {18--24 Jul},
  publisher =    {PMLR},
  url = 	 {https://proceedings.mlr.press/v139/ramesh21a.html}
}

@inproceedings{panfusion2024,
  title={Taming Stable Diffusion for Text to 360$^\circ$ Panorama Image Generation},
  author={Zhang, Cheng and Wu, Qianyi and Cruz Gambardella, Camilo and Huang, Xiaoshui and Phung, Dinh and Ouyang, Wanli and Cai, Jianfei},
  booktitle={Proceedings of the IEEE/CVF Conference on Computer Vision and Pattern Recognition},
  year={2024}
}

@inproceedings{MVDiffusion,
 author = {Tang, Shitao and Zhang, Fuyang and Chen, Jiacheng and Wang, Peng and Furukawa, Yasutaka},
 booktitle = {Advances in Neural Information Processing Systems},
 doi = {10.52202/075280-2229},
 editor = {A. Oh and T. Naumann and A. Globerson and K. Saenko and M. Hardt and S. Levine},
 pages = {51202--51233},
 publisher = {Curran Associates, Inc.},
 title = {MVDiffusion: Enabling Holistic Multi-view Image Generation with Correspondence-Aware Diffusion},
 url = {https://proceedings.neurips.cc/paper_files/paper/2023/file/a0da690a47b2f52faa63f6fe054057b5-Paper-Conference.pdf},
 volume = {36},
 year = {2023}
}

@inproceedings{Lee2023SyncDiffusionCM,
 author = {Lee, Yuseung and Kim, Kunho and Kim, Hyunjin and Sung, Minhyuk},
 booktitle = {Advances in Neural Information Processing Systems},
 editor = {A. Oh and T. Naumann and A. Globerson and K. Saenko and M. Hardt and S. Levine},
 pages = {50648--50660},
 publisher = {Curran Associates, Inc.},
 title = {SyncDiffusion: Coherent Montage via Synchronized Joint Diffusions},
 url = {https://proceedings.neurips.cc/paper_files/paper/2023/file/9ee3a664ccfeabc0da16ac6f1f1cfe59-Paper-Conference.pdf},
 volume = {36},
 year = {2023}
}

@inproceedings{BarTal2023MultiDiffusionFD,
  title={MultiDiffusion: Fusing Diffusion Paths for Controlled Image Generation},
  author={Omer Bar-Tal and Lior Yariv and Yaron Lipman and Tali Dekel},
  booktitle={International Conference on Machine Learning},
  year={2023},
  url={https://api.semanticscholar.org/CorpusID:256900756}
}

@inproceedings{wang2023360,
  title={360-Degree Panorama Generation from Few Unregistered NFoV Images},
  author={Wang, Jionghao and Chen, Ziyu and Ling, Jun and Xie, Rong and Song, Li},
  booktitle={Proceedings of the 31st ACM International Conference on Multimedia},
  pages={6811--6821},
  year={2023}
}

@inproceedings{DiffusionBeatGAN,
 author = {Dhariwal, Prafulla and Nichol, Alexander},
 booktitle = {Advances in Neural Information Processing Systems},
 editor = {M. Ranzato and A. Beygelzimer and Y. Dauphin and P.S. Liang and J. Wortman Vaughan},
 pages = {8780--8794},
 publisher = {Curran Associates, Inc.},
 title = {Diffusion Models Beat GANs on Image Synthesis},
 url = {https://proceedings.neurips.cc/paper_files/paper/2021/file/49ad23d1ec9fa4bd8d77d02681df5cfa-Paper.pdf},
 volume = {34},
 year = {2021}
}

@inproceedings{ddpm,
 author = {Ho, Jonathan and Jain, Ajay and Abbeel, Pieter},
 booktitle = {Advances in Neural Information Processing Systems},
 editor = {H. Larochelle and M. Ranzato and R. Hadsell and M.F. Balcan and H. Lin},
 pages = {6840--6851},
 publisher = {Curran Associates, Inc.},
 title = {Denoising Diffusion Probabilistic Models},
 url = {https://proceedings.neurips.cc/paper_files/paper/2020/file/4c5bcfec8584af0d967f1ab10179ca4b-Paper.pdf},
 volume = {33},
 year = {2020}
}

@article{chen2022text2light,
    title={Text2Light: Zero-Shot Text-Driven HDR Panorama Generation},
    author={Chen, Zhaoxi and Wang, Guangcong and Liu, Ziwei},
    journal={ACM Transactions on Graphics (TOG)},
    volume={41},
    number={6},
    articleno={195},
    pages={1--16},
    year={2022},
    publisher={ACM New York, NY, USA}
}

\end{document}